\documentclass[conference]{IEEEtran}
\IEEEoverridecommandlockouts
\usepackage{cite}
\usepackage{amsmath,amssymb,amsfonts}
\usepackage{adjustbox}
\usepackage{algorithmic}
\usepackage{graphicx}
\usepackage{textcomp}
\usepackage{float}
\newcommand{\etal}{\textit{et al.}}
\usepackage{xcolor}
\def\BibTeX{{\rm B\kern-.05em{\sc i\kern-.025em b}\kern-.08em
    T\kern-.1667em\lower.7ex\hbox{E}\kern-.125emX}}
\begin{document}

\title{Pedestrian Behavior Maps for Safety Advisories: \\ CHAMP Framework and Real-World Data Analysis}

\author{\IEEEauthorblockN{Ross Greer* \thanks{*All authors are associated with the Laboratory for Intelligent \& Safe Automobiles (LISA) at the University of California, San Diego in La Jolla, USA. All authors contributed equally. }}
\IEEEauthorblockA{
regreer@ucsd.edu}
\and
\IEEEauthorblockN{Samveed Desai}
\IEEEauthorblockA{
s7desai@ucsd.edu}
\and
\IEEEauthorblockN{Lulua Rakla}
\IEEEauthorblockA{
lrakla@ucsd.edu}
\and
\IEEEauthorblockN{Akshay Gopalkrishnan}
\IEEEauthorblockA{
agopalkr@ucsd.edu}
\and
\IEEEauthorblockN{Afnan Alofi}
\IEEEauthorblockA{
aalofi@ucsd.edu}
\and
\IEEEauthorblockN{Mohan Trivedi}
\IEEEauthorblockA{
mtrivedi@ucsd.edu}
}

\maketitle

\begin{abstract}
It is critical for vehicles to prevent any collisions with pedestrians. Current methods for pedestrian collision prevention focus on integrating visual pedestrian detectors with Automatic Emergency Braking (AEB) systems which can trigger warnings and apply brakes as a pedestrian enters a vehicle’s path. Unfortunately, pedestrian-detection-based systems can be hindered in certain situations such as night-time or when pedestrians are occluded. Our system addresses such issues using an online, map-based pedestrian detection aggregation system where common pedestrian locations are learned after repeated passes of locations. Using a carefully collected and annotated dataset in La Jolla, CA, we demonstrate the system's ability to learn pedestrian zones and generate advisory notices when a vehicle is approaching a pedestrian despite challenges like dark lighting or pedestrian occlusion. Using the number of correct advisories, false advisories, and missed advisories to define precision and recall performance metrics, we evaluate our system and discuss future positive effects with further data collection. We have made our code available at https://github.com/s7desai/ped-mapping, and a video demonstration of the CHAMP system at https://youtu.be/dxeCrS\_Gpkw.
\end{abstract}


\section{Introduction}

According to a report from the National Highway Traffic Safety Administration \cite{1}, 6,516 pedestrians died and 54,769 were injured in traffic crashes in the United States in 2020. From 2000 to 2020, there has been an increase by 42\% in the number of pedestrian fatalities on public roadways despite developments in vehicle and road safety. The majority of pedestrian traffic deaths occurred in urban areas (80\%) and on open roads (75\%) rather than at intersections (25\%). The report data showed that the primary factor of pedestrian traffic deaths (50\%) was the failure to yield right of way. Due to the prioritization of infrastructure designed for the convenience of cars over the last few decades, safe and convenient pedestrian infrastructure has been dramatically reduced. As a result, outside of urban areas with high walk-scores, pedestrian activity is sparsely distributed. However, in most cases, pedestrian patterns emerge that may or may not align with existing pedestrian paths or marked crossings. In these areas, it can be difficult for new drivers, or drivers unfamiliar with certain neighborhoods to understand where to expect pedestrians. 

For current pedestrian detectors, vehicles often use cameras or radar to check if there are pedestrians ahead of them. If a pedestrian is in the car’s path, then a warning will be triggered by the vehicle to alert the driver. These warnings can vary from something like a popup message on a dashboard, alert sound, or a seat vibration. These pedestrian detectors and alert systems can be combined with Automatic Emergency Braking (AEB) to avoid colliding with a pedestrian.

\begin{figure}
    \centering
    \includegraphics[width=.25\textwidth]{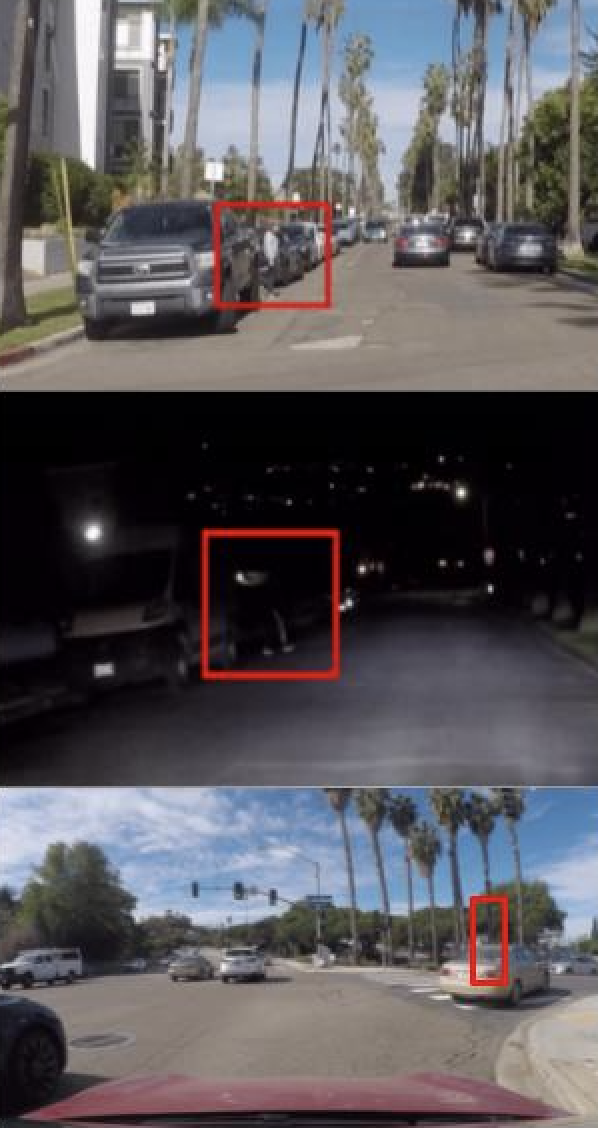}
    \caption{Our system, CHAMP, aims to issue advisories in safety critical scenarios like occluded pedestrians (top left), dark lighting conditions (top right) and blind right turns (bottom). Pedestrian bounding boxes drawn for illustrative purposes. }
\label{fig1}
\end{figure}

Automatic Emergency Braking (AEB) systems have improved pedestrian safety by reducing vehicle velocity to avoid or mitigate effects of collision, an ongoing risk when pedestrians are present at both marked and unmarked crossings (\cite{2,3,4,5}) and intersections. AEB can be performed through the use of something like a PD controller which can stabilize the vehicle's velocity and bring it to a stop \cite{nguyen2022pedestrian}. However, Haus \etal \cite{2} show that all AEB models in their study were actually more effective when combined with driver braking, as opposed to autonomous behavior alone. Further, despite their effectiveness in certain situations, AEB systems are less effective in common driving scenarios. Cicchino \cite{3} and a recent AAA study \cite{4} show evidence that AEB systems are less effective in dark conditions, when vehicles are turning, and when pedestrians appear suddenly from occluded locations - all very common scenarios for pedestrian encounters. In such situations, vision-based pedestrian detectors are not as effective because these situations lack visual cues for the vehicle to recognize an approaching pedestrian. As Sohail \etal \cite{sohail2023data} show, pedestrian safety is dependent on demographics with young and old people being at more risk. 

To address this performance gap and implement these motivating principles, we propose a system that advises the driver when they are in the likely presence of pedestrians. We name this system CHAMP, which stands for \textit{Crowd-sourced, History-Based Advisories of Mapped Pedestrians}. Rather than being based on single-instance detections, the proposed system is an online, map-based approach, where pedestrian behaviors are aggregated and learned from repeated passes of vehicles with at least a single camera. The repetition of visits mitigates the effects of detection failure cases and environmental changes (lighting, occlusion, vehicle path). From the generated map, statistical pedestrian aggregation or crossing patterns are inferred. Finally, an advisory threshold is created which can be tailored to driver preference or safety standard, with a possible range from activation if a pedestrian has ever crossed in the area, to activation only in areas where pedestrian activity matches or exceeds the flow of a high-traffic signalized crosswalk.  

Our proposed mapping and advisory system, which to the best of our knowledge is the first of its kind at the time of writing, allows multiple benefit opportunities: 
\begin{itemize}
    \item advisories for driver vigilance in safety-critical pedestrian situations, 
    \item earlier recommendations for braking readiness in domains where AEB systems may fail or similar ADAS may benefit from driver cooperation, such as night-time conditions or driving around occluded pedestrians, and
    \item development of informative HD map layers which provide an estimate of notable prior pedestrian activity, which can be thresholded according to the goals of an autonomous driving or driver assistance system. 
\end{itemize}

\section{Related Research}

Current studies that explore pedestrian detection can be categorized in several ways. First, some studies focus solely on creating pedestrian detection datasets, whether the focus be on a large dataset that reflects a variety of conditions involving pedestrians \cite{6} or a focus solely on pedestrian detection in monocular images in urban environments \cite{7}; however, a weakness of these datasets is that not all of them are entirely human-annotated. Some of these pedestrian detection datasets are a subset of a larger traffic object dataset which includes annotations of surrounding vehicles, signs, pedestrians, \emph{etc}. \cite{cress2022a9} \cite{greer2022salience} \cite{greer2023salient}. Second, other studies can be categorized by the conditions in which they attempt to detect and classify pedestrians, such as studies that use deep learning techniques and CNNs to detect pedestrians specifically in hazy weather conditions \cite{8} and a variety of weather conditions \cite{9}, studies that use SVM for detection and Kalman filters for tracking of pedestrians in low-visibility conditions at night\cite{10}, and studies that use motion estimation and texture analysis techniques in order to detect pedestrians in crowded urban environments specifically \cite{11}; however, many of these studies solely focus on detection of existing pedestrians, rather than predicting the likely movement trajectories of pedestrians. Some pedestrian detection works focus on crowd-sourced pedestrian hotspots rather than focusing on single detections and predictions. Typically, this is in the form of detecting vehicle-pedestrian collision hotspots using crowdsourced pedestrian data. To examine pedestrian safety in urban areas, Telima \emph{et al}. use crowdsourced traffic incident data to predict pedestrian hotspot locations during accidents \cite{telima2023use}. Yao \emph{et al}. perform a similar task by identifying vehicle-pedestrian collision hotspots with a network kernel density estimator and using this to train a random forest to model the observed vehicle-pedestrian crash densities \cite{yao2018identification}.

Predicting the movement and trajectory of pedestrians is another common research task toward safe driving systems \cite{mogelmose2015trajectory}. As described by Ridel et al. \cite{ridel2018literature}, estimating pedestrian motion is difficult because pedestrians move with random movements and can be occluded by various objects. Estimating pedestrian motion is similar to a vehicle trajectory prediction task \cite{deo2020trajectory} \cite{messaoud2021trajectory} and therefore can borrow ideas from this research. For example, using a birds-eye-roadmap for trajectory predictions \cite{greer2021trajectory} would be beneficial to pedestrian trajectory estimation models.  Research on predicting the movement of pedestrians can be categorized based on the techniques that they employ to predict these trajectories:
\begin{itemize}
    \item Asahara et al.\cite{12} use a mixed Markov-chain model,
    \item Particke et al. \cite{13} use an advanced Kalman filter called a Multi-Hypotheses filter,
    \item Song et al. \cite{14} use tensors to represent features of pedestrians in crowded environments and a convolutional LSTM,
    \item Ridel et al. use LSTM-based trajectory prediction \cite{ridel2019understanding} using not only past pedestrian trajectory, but also ego vehicle trajectory and pedestrian heading, 
    \item Keller et al.\cite{15} use stereo-vision based path prediction,
    \item Syed and Morris \cite{syed2021stgt} model inter-pedestrian behavior through a graph structure of and then predict pedestrian motion through a Transformer network,
    \item Alemaw et al. \cite{alemaw2022data} estimate the states of interacting agents like pedestrians by fusing odometry and video data in a Multi-Agent Hierarchical Dynamic Bayesian Network (MAH-DBN) framework, and
    \item Ridel et al. \cite{ridel2020scene} estimate long term human motion through a model by representing the trajectories of agents with a binary 2D grid and the scene as a birds-eye view image.
\end{itemize}
    However, some of these studies only have a limited number of pedestrian movements that they consider; for instance, Keller et al. \cite{15} only consider pedestrian movement lateral to the vehicle where the pedestrian stops and when the pedestrian continues to walk past the vehicle. A survey of methods used to predict pedestrian behavior \cite{ridel2018literature} shows that most systems use active detections using LiDAR, stereo, or thermal cameras. But, these systems provide a short response time to the driver and fail when the detector fails to identify the pedestrian. By contrast, our system CHAMP employs a method of aggregating pedestrian behaviors after repeated detections and passes. 

\section{CHAMP Computational Framework}

\subsection{Aggregation Methods: Fleet and Repeat}
Our work addresses the problem of safe pedestrian zone mapping through two principles, which we name “Fleet” and “Repeat”. These terms signify that long-term patterns in scene agent behavior can be uncovered using repeated samplings of a geographic location, either in a spatial sense (e.g. covered by a fleet of vehicles, “Fleet”) or temporal sense (e.g. repeated visits from a single car, “Repeat”). 

\begin{figure}
    \includegraphics[width=.5\textwidth]{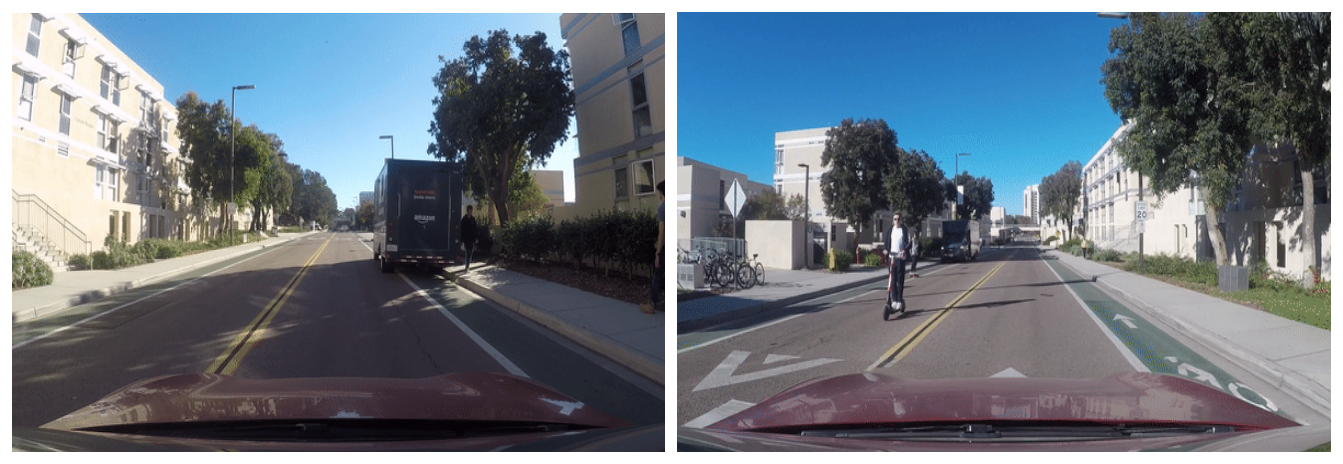}
    \caption{A large truck occludes a view of pedestrians from either side of the truck, but having a vehicle positioned on both ends of the truck would capture enough information to remove this blind spot in a spatial sense. Similarly, returning to capture a new image after the truck has left the area would allow for resolution of the blind spot in a temporal sense. Such temporal blind spots may also be induced by poor lighting. The need to resolve these blind spots motivates a method of “Fleet or Repeat”, whereby repeatedly sampling scenes we can detect pedestrians occluded from specific single views.}
    \label{fig2}
\end{figure}

\begin{figure}
    \includegraphics[width=.5\textwidth]{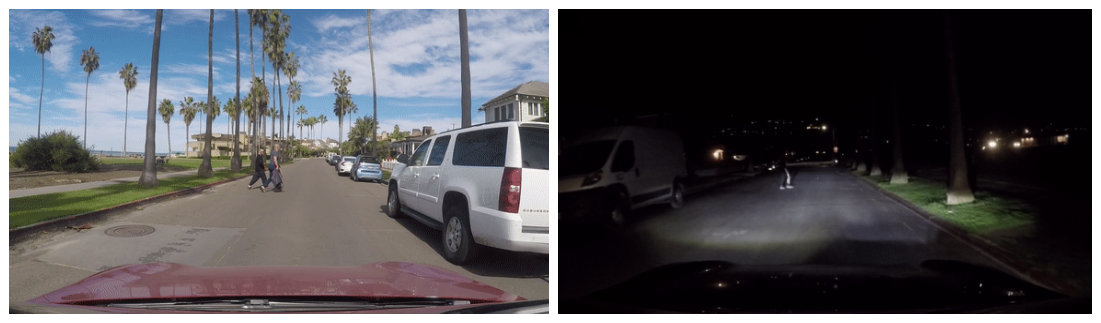}
    \caption{During the day, several pedestrians are spotted on a stretch of the road. During the night, when the vehicle drives along the same road, given the poor lighting conditions, it would be difficult to catch blind spots. But, using the drive during the day allows for blind spots to be concealed in a temporal sense. The “Fleet or Repeat” method can hereby sample scenes with pedestrians during the day and help detect occluded pedestrians during the night.}
    \label{fig3}
\end{figure}

As two examples of this method of repeated sampling to infer real-world patterns, Morris and Trivedi \cite{16} show that repeated polls of detected vehicles on roadways and intersections over time yield clear lane patterns and turning flows, without any sampling of the actual infrastructure. Greer and Trivedi \cite{17} show that repeated polls of a pedestrian intersection allow to map and learn clear crosswalks without sampling the infrastructure itself. In principle, these long term-data allows systems to learn behavioral patterns of pedestrians and how they interact with road infrastructure, and both methods make use only of detected dynamic scene agents (vehicles and people) to infer behavioral patterns. Further, repeated visits also provide a secondary view of an area that may have been a blind spot (either in direction or in time), as illustrated in Figures \ref{fig2} and \ref{fig3}. 

\subsection{System Design}
\begin{figure*}
    \includegraphics[width=\textwidth]{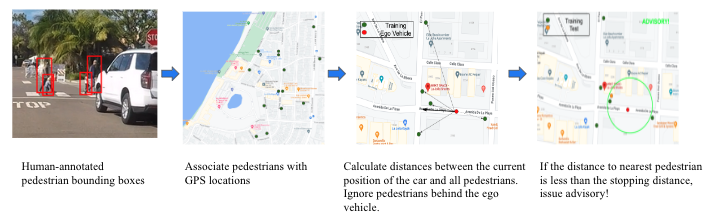}
    \caption{System design of CHAMP, with three stages: association of pedestrian detections to a vehicle GPS location, nearest pedestrian hotspot search, and advisory issuance. Here, pedestrians seen during training drives are represented as red dots, ego vehicle as the blue dot and pedestrian seen during test drive as a orange dot. Note that in our implementation, we utilize human-annotated detections, but in scaling such a system to operate on a fleet and update in real-time, such annotation may be done by a robust automated 2D object detector.}
    \label{fig4}
\end{figure*}

Following data collection, the CHAMP system consists of 3 components as illustrated in Figure \ref{fig4}, corresponding to training, inference, and interaction: 

\begin{itemize}
\item \textbf{Training}: Pedestrian Location Association
\item \textbf{Inference}: Nearest Pedestrian Hotspots Search
\item \textbf{Interaction}: Advisory Issuance
\end{itemize}

\subsubsection{Training: Pedestrian Location Association}
CHAMP takes in a vehicle's GPS information, in our case, recorded at a rate of up to 10 Hz, from the drive training data collection and splits it into intervals of 1 second each, a sufficient sampling frequency for the expected speed a car would have around pedestrian hotspots. Further, since we do repeated sampling of areas, the probability of missed detection of some pedestrian in a pedestrian-dense area reduces. This aggregation is done across all frames of the drive data captured by the vehicle. Using the timestamps of each GPS sample, we record $p$ location coordinates within the 1-second interval associated with a given time $t$: 

\begin{equation}
    I_t = \{(x_1,y_1), (x_2,y_2),...,(x_p,y_p)\}
\end{equation}

CHAMP then takes the median of the $p$ location coordinates of the vehicle associated with the $k$th interval $I_k$, and associates to this position the number of pedestrians $c_k$ which were detected in $I_k$:

\begin{equation}
    (\text{median}(I_k), c_k)
\end{equation}



This algorithm continually adds new pedestrian aggregation nodes as further area is covered and data is collected, scaling with spatial spread and coverage over time in accordance with the principles of ``Fleet" and ``Repeat" introduced earlier. 

\subsubsection{Inference: Finding Nearest Pedestrian Hotspots}
At inference time, our algorithm uses a ball tree nearest neighbor search algorithm \cite{moore2003new}, with haversine distance as the metric, to efficiently calculate the distance between the car position and the nearest pedestrian location $d_{\text{car,ped}}$. This approach uses a binary tree with a hierarchical structure to effectively compute the nearest neighbor between the current car position and all the pedestrians in a multidimensional space, thus reducing the overall latency during test time (from O(n) to O(log(n)) ). Our algorithm also checks if the pedestrian is in the field of view of the car by calculating the heading angle $\Theta_{heading}$ between the current position of the ego vehicle and the nearest pedestrian, so that it can ignore pedestrians behind the vehicle.

\subsubsection{Interaction: Issuing Advisory}
CHAMP then uses stopping distance\cite{18}, which incorporates velocity as a proxy to Time-To-Collision (TTC), to calculate the radius around the car to check for pedestrians. To allow for additional safety, we also use a multiplicative offset, \emph{b}. Our extended stopping distance \emph{s} is  defined as

\begin{equation}
s = b\cdot ({0.278\cdot t \cdot v + \frac{v^2}{254(f + G)})}
\end{equation} 
where \emph{t} is reaction time, \emph{v} is the velocity of the car (km/h), \emph{f} is the coefficient of friction, and \emph{G} is the slope. For our calculations, we used the approximated values \emph{f} = 0.7 (assuming a dry road) and \emph{G} = 0 (assuming most of our roads are flat without any uphill/downhill). This stopping distance acts as the advisory threshold and it changes depending on the velocity of the ego vehicle. This formula is used from the book, ``A Policy on Geometric Design of Highways and Streets" released by AASHTO (American Association of State Highway and Transportation Officials)\cite{transportation2011policy}. We are also using the reaction time \emph{t} = 2.5 s, as this emcompasses the reaction times of most drivers, including elderly drivers.

To reduce repetitive computations, CHAMP performs all the above calculations every $K$ meters (where $K$ is the \textit{sampling distance}). $K$ is a hyperparameter that we can tune according to observed performance considering vehicle speeds, GPS sampling rates, and pedestrian frequency.

During inference, if the distance between the car and the nearest pedestrian is less than the stopping distance ($d_{\text{car, ped}} \leq s$) and the heading angle is less than 90 degrees ($\Theta_{\text{heading}} \leq 90$), CHAMP issues an advisory to the driver. These advisories request the driver to be vigilant and keep their eyes on the road, making the driver aware that pedestrians may be present.

\section{System Evaluation}
We gathered real-world driving data in the La Jolla area of San Diego, California area. Data was captured by a front-facing GoPro camera and GPS sensor mounted to the LISA-T testbed \cite{19}. We aggregated 10,000 clips of 10 seconds each, which at a frame rate of 30 FPS resulted in 3 million stored frames. We restricted the data to the La Jolla region of San Diego to facilitate testing and training with best possible coverage under our training data volume. After collection, expert human annotators marked keyframes (sampled at 1 Hz), in particular annotating 2D bounding boxes around pedestrians among other road objects using AWS Sagemaker. We have associated the locations of the pedestrians with the GPS locations of the ego vehicle at that particular time instant. The dataset was stored in a csv format and contained the following information: timestamp, latitude, longitude, pedestrian count and the clipID, corresponding to which clip the GPS information was associating to. Important to this task, clips may feature pedestrians crossing in the vicinity of the ego vehicle, and these pedestrians are specifically annotated as exhibiting crossing behavior. While this is done manually for the purposes of initial evaluation, machine learning techniques for goal-oriented multi-agent tracking may further reduce the need for human annotation on such subtasks. 

In addition to training data collection, we took an additional independent set of drives to generate test clips for evaluation of CHAMP, returning to previously visited areas at different times and from different directions. These drives captured typical scenarios as well as scenarios including blind turns, occluded areas, and dark lighting conditions. CHAMP was then used to process the test data, producing advisory alerts at different locations. These advisory periods were then compared against human-defined ground truth scenarios to demonstrate the system quantitatively and qualitatively, discussed in the following sections.  In the current scheme, the entire past history (all training data) is used to issue advisories. This can be changed to make the system more sensitive to recent pedestrian history. 

For this and subsequent sections, we define training as aggregating vehicle GPS locations from where pedestrians are detected (a method by which we represent the pedestrian behavior as a latent variable) and testing as evaluating system performance using data from test drives. This is different from the term training used in machine learning, where a model is fit on training data.

\subsection{Quantitative Analysis}
CHAMP is evaluated using standard precision and recall metrics. Precision is the ratio of the correct advisories to the total advisories given, used to quantify the prevalence of false advisories. Recall is the ratio of the correct advisories to the total advisories the system should have given, used to quantify missed advisories. Both metrics are described by the equations: 
\begin{equation}
\resizebox{.7\hsize}{!}{$\text{Precision} = \frac{|\text{Correct Advisories}|}{|\text{Correct Advisories}| + |\text{False Advisories}|}$}
\end{equation}

\begin{equation}
\resizebox{.7\hsize}{!}{$\text{Recall} = \frac{|\text{Correct Advisories}|}{|\text{Correct Advisories}| + |\text{Missed Advisories}|}$}
\end{equation}

As CHAMP is intended to be a safety system, recall is a prioritized metric, as it is critical to be vigilant in zones where pedestrian behavior is prevalent and avoid missed advisories (i.e. preference for over-advising to under-advising). There is a natural trade-off between recall and precision (over-advising and under-advising)  as low precision can cause the driver to get distracted with false positives and might cause them to turn off the advisories. We explain strategies for mitigating over-advising later in this section.

Another hyperparameter that can be modified to influence system performance is sampling distance $K$, the distance traveled between consecutive assessments for advisory. This controls how often we compute the nearest pedestrian from the current location of the car. From our observations, early detection of the nearest pedestrian can be achieved with a minimal sampling distance of 2 meters. On reducing the sampling distance further, we noticed that the precision drops significantly as there are a lot of false positives which show up in the sample. Sampling at lower frequency allows for reasonably reduced computational cost, and the effects of sampling distance on performance are displayed in Tables \ref{table:precision1} to \ref{table:precision5}.

\begin{figure}[!htb]
    \centering
    \includegraphics[width=.45\textwidth]{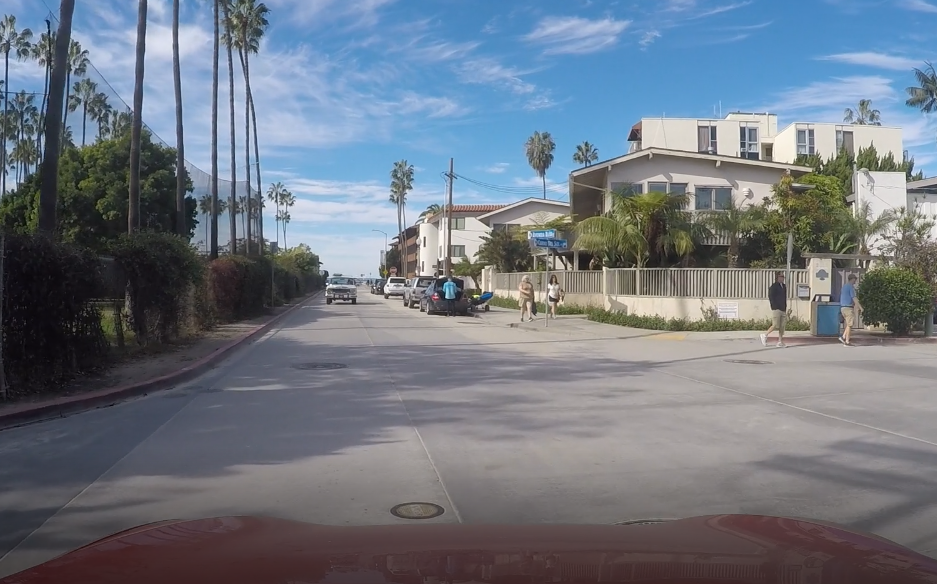}
    \caption{Image of Test Clip ID 4 (Densely Populated Area)}
    \label{fig5}
\end{figure}

\begin{figure}[!htb]
    \centering
    \includegraphics[width=.45\textwidth]{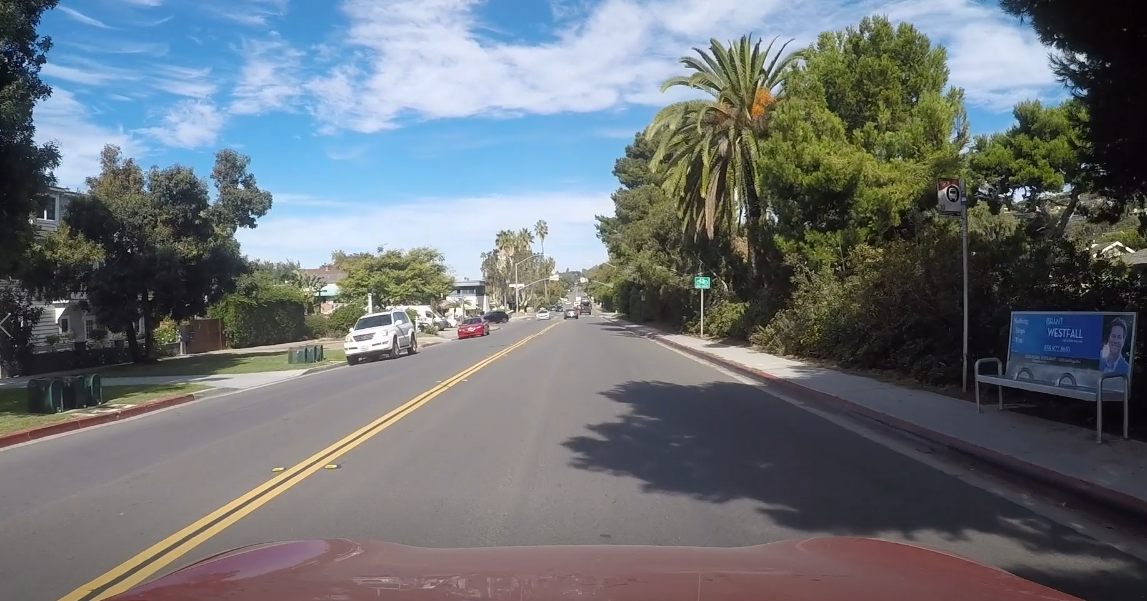}
    \caption{Image of Test Clip ID 5 (Infrequently Used Intersection)}
    \label{fig6}
\end{figure}

Clips selected for the test set include coverage of general areas previously visited at different times during training, as well as locations with environments similar to typical failure cases for pedestrian detection or AEB systems as described in related research. The example locations analyzed in our test set are as follows:
\begin{itemize}
    \item \textbf{Clip 1} (Figure \ref{fig9}) contains a blind right turn.
    \item \textbf{Clip 2} (Figures \ref{fig7}, \ref{fig8}) contains pedestrians occluded by vehicles before crossing.
    \item \textbf{Clip 3} (Figure \ref{fig3}) contains dark, night-time conditions.
    \item \textbf{Clip 4} (Figure \ref{fig5}) contains a densely populated intersection area.
    \item \textbf{Clip 5} (Figure \ref{fig6}) contains an infrequently-crossed intersection.
\end{itemize}

In our examples, we find the expected result that lower sampling distances generally have better recall, as certain pedestrians may be missed within higher spans between computations (such as our largest 5 meter sampling distance). Hence, we suggest a sampling distance of 2 meters (though this can be tuned further depending on the driving environment) to balance precision and recall. 

Unique qualities to some scenarios which affect system performance are explained here:
\begin{itemize}
    \item Pedestrian detections are so saturated in Clip 4 that sampling distance has little effect on whether an advisory is issued; at any distance, it is likely that a crossing pedestrian has been observed in the training data, so for this locale, advisories are issued frequently. This increases the number of false positives, thereby lowering precision.
    \item Clip 5 represents a case with minimal sampling (two passes from the training vehicle), and in each case, one non-crossing pedestrian was detected. Because no actively crossing pedestrians were observed in training, the system has poor (0) recall at test. This would be resolved with increased data collection; presumably, pedestrians would be found crossing if the location were polled additional times. 
\end{itemize}

While an expanded dataset would be the most helpful component to the system, to further increase precision and reduce false advisories we also propose future research in denoising and thresholding steps, including:
\begin{enumerate}
    \item Increased sampling to allow for filtering by frequency of pedestrian sightings in a given area. 
    \item Incorporation of trajectory prediction models so that pedestrian data can be automatically filtered to include only instances where pedestrians actually cross the street without excessive human annotation. With this additional data available, the system can be relieved of false positives associated with a crowded pedestrian scene in which the pedestrians are consistently compliant with sidewalk and non-roadway trajectories. Machine learning systems which analyze for pedestrian intention \cite{fang2017board} \cite{volz2016data} would be useful toward this end. 
    \item Stratification of advisories by time of the day (e.g. a school zone may issue advisories around the beginning or end of the school day, or a street with busy nightlife may issue advisories during night-time drives). Such time-of-day information is currently collected as meta-data, and under higher volumes of data collection, will provide the means for issuing advisories specific to the time-of-day patterns in pedestrian occupancy. 
\end{enumerate}


\begin{table}[h!]
\centering
\caption{Precision and Recall by Sampling Distance \\ Test Clip ID 1 (Blind Turn)}
\label{table:precision1}
\begin{adjustbox}{max width=0.8\textwidth}
\begin{tabular}{|c|c|c|} 
\hline
\begin{tabular}[c]{@{}c@{}}\textbf{Sampling }\\\textbf{Distance (m)}\end{tabular} & \textbf{Precision} & \textbf{Recall}  \\
\hline
\textbf{2}                   & \textbf{0.27} & \textbf{0.75}                       \\ 
\hline
3                   & 0.4                    & 0.5                        \\ 
\hline
4                   & 1                      & 0.25                       \\ 
\hline
5                   & 1                      & 0.25                       \\ 
\hline
\end{tabular}
\end{adjustbox}
\end{table}

\begin{table}[h!]
\centering
\caption{Precision and Recall by Sampling Distance \\ Test Clip ID 2 (Occluded Pedestrians)}
\label{table:precision2}
\begin{adjustbox}{max width=0.8\textwidth}
\begin{tabular}{|c|c|c|} 
\hline
\begin{tabular}[c]{@{}c@{}}\textbf{Sampling }\\\textbf{Distance (m)}\end{tabular} & \textbf{Precision} & \textbf{Recall}  \\
\hline
\textbf{2}          & \textbf{0.75}         & \textbf{0.75}                       \\ 
\hline
3                   & 0.43                   & 0.6                        \\ 
\hline
4                   & 0.6                   & 0.4                       \\ 
\hline
5                   & 1                      & 0.25                       \\ 
\hline
\end{tabular}
\end{adjustbox}
\end{table}

\begin{table}[h!]
\centering
\caption{Precision and Recall by Sampling Distance \\ Test Clip ID 3 (Night-time Lighting)}
\label{table:precision3}
\begin{adjustbox}{max width=0.8\textwidth}
\begin{tabular}{|c|c|c|} 
\hline
\begin{tabular}[c]{@{}c@{}}\textbf{Sampling }\\\textbf{Distance (m)}\end{tabular} & \textbf{Precision} & \textbf{Recall}  \\
\hline
\textbf{2}          & \textbf{0.105}         & \textbf{0.72}                       \\ 
\hline
3                   & 0.3                    & 0.6                        \\ 
\hline
4                   & 0.75                   & 0.4                       \\ 
\hline
5                   & 0.75                   & 0.4                       \\ 
\hline
\end{tabular}
\end{adjustbox}
\end{table}

\begin{table}[h!]
\centering
\caption{Precision and Recall by Sampling Distance \\ Test Clip ID 4 (Densely Populated Area)}
\label{table:precision4}
\begin{adjustbox}{max width=0.8\textwidth}
\begin{tabular}{|c|c|c|} 
\hline
\begin{tabular}[c]{@{}c@{}}\textbf{Sampling }\\\textbf{Distance (m)}\end{tabular} & \textbf{Precision} & \textbf{Recall}  \\
\hline
\textbf{2}          & \textbf{0.16}          & \textbf{0.57}                       \\ 
\hline
3                   & 0.18                   & 0.5                        \\ 
\hline
4                   & 0.18                   & 0.5                       \\ 
\hline
5                   & 0.18                   & 0.43                       \\ 
\hline
\end{tabular}
\end{adjustbox}
\end{table}

\begin{table}[h!]
\centering
\caption{Precision and Recall by Sampling Distance \\ Test Clip ID 5 (Infrequently used intersection)}
\label{table:precision5}
\begin{adjustbox}{max width=0.8\textwidth}
\begin{tabular}{|c|c|c|} 
\hline
\begin{tabular}[c]{@{}c@{}}\textbf{Sampling }\\\textbf{Distance (m)}\end{tabular} & \textbf{Precision} & \textbf{Recall}  \\
\hline
2                   & 0                       & 0                       \\ 
\hline
3                   & 0                       & 0                        \\ 
\hline
4                   & 0                       & 0                       \\ 
\hline
5                   & 0                       & 0                       \\ 
\hline
\end{tabular}
\end{adjustbox}
\end{table}

\subsection{Qualitative Analysis}

\begin{figure}
    \includegraphics[width=.5\textwidth]{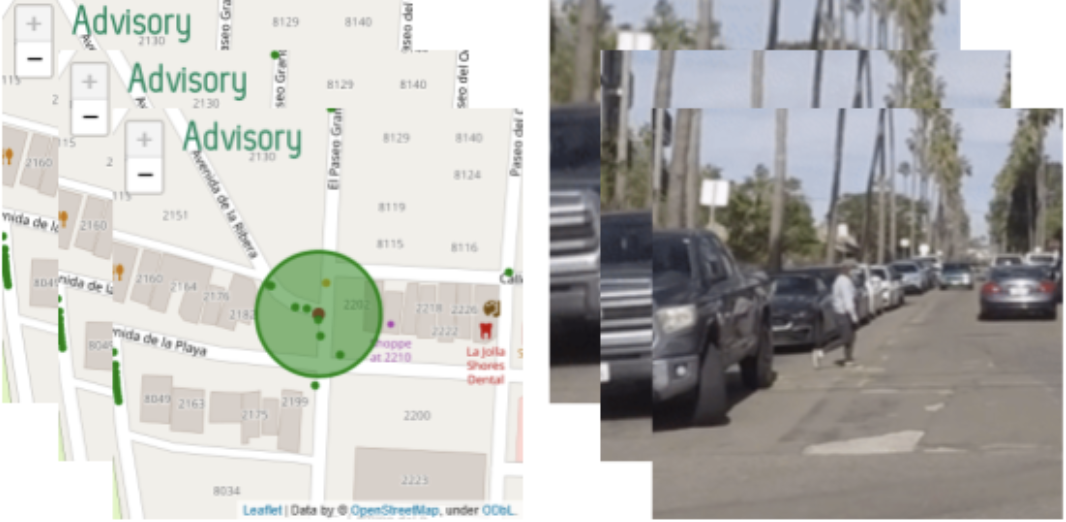}
    \caption{CHAMP issues a warning in sequential frames to advise the driver that there may be pedestrians nearby, based on the training data. In the image on the left, the red dot represents the ego-vehicle, the green dots are pedestrians seen in the training set, and the orange dot is the unseen pedestrian in the test data.}
    \label{fig7}
\end{figure}

Figure \ref{fig7} illustrates a scene from Clip 2 in the test data, featuring pedestrians emerging from an occluded area. During training, CHAMP has seen pedestrians in this area that are within the radius of the advisory distance and in front of the car, so CHAMP issues a vigilance advisory notice to the driver. This allows for the driver to be alert for the occluded pedestrian crossing the street. 

\begin{figure}
    \includegraphics[width=0.5\textwidth]{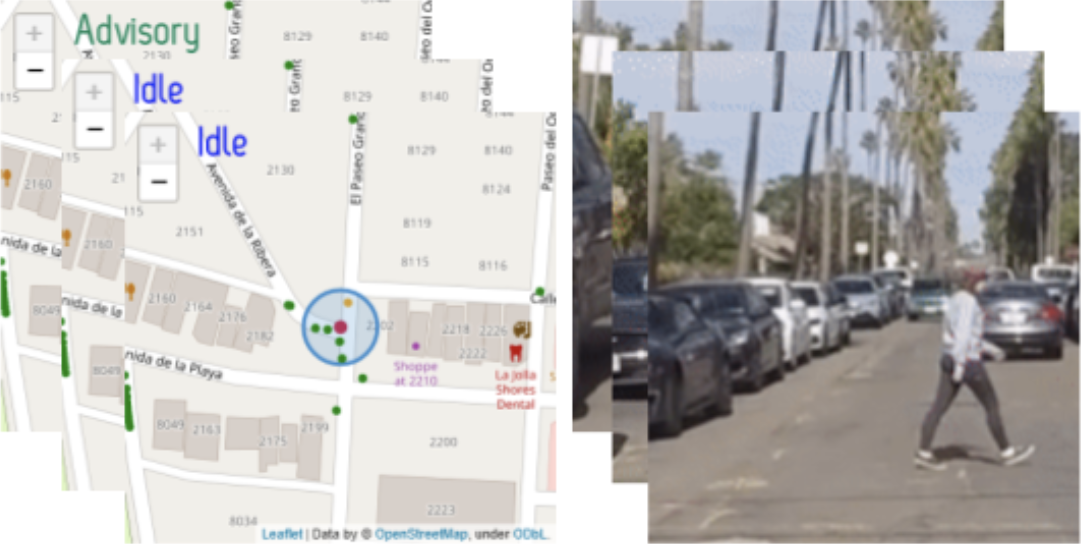}
    \caption{CHAMP turns off advisory mode after passing all of the pedestrian training points.}
    \label{fig8}
\end{figure}

Now that the driver has been advised by CHAMP, they can properly slow down and wait for the pedestrian to cross the street. In Figure \ref{fig8}, we can see that once we are behind any of the training points the vigilance advisory notice will turn off. The key point to note here is that CHAMP tries to issue an advisory before the car enters the hotspot, and is designed to turn off when the car is already in the hotspot. It is a preemptive advisory.

\begin{figure}
    \centering
    \includegraphics[width=0.5\textwidth]{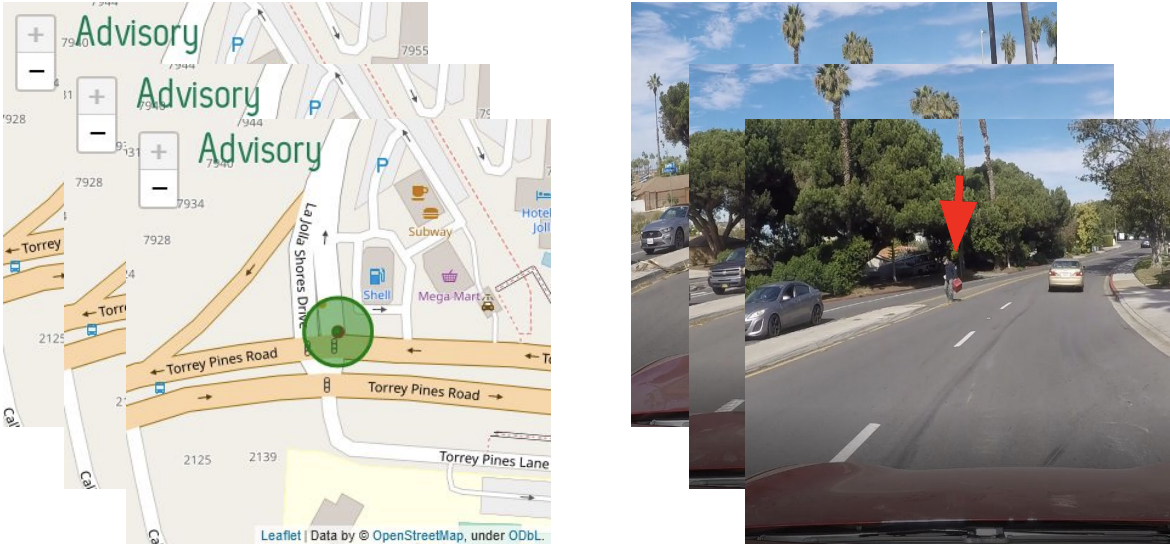}
    \caption{CHAMP maintains active advisory mode as a pedestrian crosses the street where there is no crosswalk.}
    \label{fig9}
\end{figure}

Figure \ref{fig9} also exemplifies another situation where the car is correctly advised with unseen pedestrian. After the car makes a right turn and approaches a pedestrian, CHAMP maintains advisory mode since there were pedestrians in the training data present in this region.  

A video demonstration of the CHAMP system can be found at https://youtu.be/dxeCrS\_Gpkw, showing three driving scenes accompanied by an overhead map showing training data, test data, vehicle position, CHAMP's advisory radius, and advisory state.

\section{Concluding Remarks}
\subsection{Impact on Pedestrian Safety}

Previous research reported in \cite{20} shows that Pedestrian Crash Avoidance and Mitigation (PCAM) systems can reduce crashes where the pedestrian crosses in front of the car or is along the side of the road by 77.6\%.

CHAMP’s effectiveness in dark lighting conditions and when pedestrians are occluded has the potential to prevent many pedestrian incidents. Of the total 61,285 pedestrian incidents in 2020 reported in \cite{1}, 12\% occurred due to visibility and 18\% due to improper crossing. Incidents such as these may be mitigated by CHAMP if integrated into existing navigation systems, HD maps, or automotive APIs.


\subsection{Applications and Future Research}

In continued development, there are multiple integration modalities for CHAMP: we propose applications as a map API for existing navigation apps (for example, an option within Google Maps or Waze) for drivers in manual vehicles, and integration within autonomous vehicle navigation systems. We envision autonomous cars with state-of-the-art 3D detectors to upload pedestrian hotspot locations to a CHAMP processing and distribution server to continuously update the dataset. The evaluation of detections is done by factoring the sensor suite in the autonomous car to  determine how much weight or trust should be given to that car’s input. One way to filter bad detections would be to threshold the confidence of detection e.g detections with low confidence are filtered out. Using these pedestrian locations, we can automate pedestrian data collection with a 3D annotation tool like 3D BAT \cite{zimmer20193d} and use systems like DashCam Cleaner \cite{21} to obscure the identities of detected pedestrians for privacy, enabling generation of pedestrian hotspot maps and datasets. CHAMP can then process the information, apply denoising and thresholding, and aggregate the pedestrian sightings. Included in this thresholding, for example, would be considerations on system response when the vehicle is known to be on a freeway; in such case, extra care should be taken to avoid conflating pedestrian activity on overpasses or nearby side streets with the GPS position of the vehicle. On a similar note, if a vehicle's path or driver's intention is known, this trajectory can also be used to limit advisories to off-track pedestrian hotspots to avoid further false-positive alerting. In summary, it is important for the system to be aware of and considerate towards the driving context, as a driver in a known and crowded urban area may already be safely vigilant, whereas the driver in a calmer environment where the pedestrian would be unexpected would have greater benefit (and less possible annoyance) from such an advisory; integration of these scene elements are left towards future application implementations, and similarly leave the optimal advisory mechanism (visual, auditory, or otherwise) to the current and future human interface research. 


Two benefits this system provides are (1) online detectors of autonomous cars can be leveraged and (2) any car or driver using a navigation system can benefit from CHAMP’s safety advisories. Moreover, we aim to provide some research on how noisy GPS data can affect our advisories and the maximum amount of noise that still allows CHAMP to make accurate predictions. This would require us to add random amounts of noise to our GPS readings and simulate the performance of CHAMP with this altered data. Finally, as an ancillary application from CHAMP’s maps, understanding pedestrian foot traffic can help relevant decision makers decide the road infrastructure (crosswalks, signs, \emph{etc}.). CHAMP can benefit individual drivers, autonomous vehicle performance, and urban planners by estimating and continuously updating zones of high pedestrian activity.

\section*{Acknowledgment}

The authors would like to acknowledge the support of LISA sponsors and the valuable comments from distinguished NHTSA and LISA colleagues.

\bibliography{biblio}
\bibliographystyle{IEEEtran}

\end{document}